\documentclass{article}

\usepackage[final]{neurips_2023}

\usepackage[utf8]{inputenc} 
\usepackage[T1]{fontenc}    
\usepackage{hyperref}       
\usepackage{url}            
\usepackage{booktabs}       
\usepackage{amsfonts}       
\usepackage{nicefrac}       
\usepackage{microtype}      
\usepackage{xcolor}         

\usepackage{tikz-network}
\usepackage{subcaption}
\usepackage{fontawesome}
\usepackage{multirow}

\usepackage{xcolor}

\title{MLFMF: Data Sets for Machine Learning for Mathematical Formalization}

\author{%
  Andrej Bauer \\
  Faculty of Mathematics and Physics \\
  University of Ljubljana \\
  Institute for Mathematics, Physics and Mechanics \\
  Ljubljana, Slovenia \\
  \texttt{andrej.bauer@fmf.uni-lj.si} \\
  \And
  Matej Petković \\
  Faculty of Mathematics and Physics \\
  University of Ljubljana \\
  Department of Knowledge Technologies \\
  Jožef Stefan Institute \\
  Ljubljana, Slovenia \\
  \texttt{matej.petkovic@fmf.uni-lj.si} \\
  \And
  Ljupčo Todorovski \\
  Faculty of Mathematics and Physics \\
  University of Ljubljana \\
  Department of Knowledge Technologies \\
  Jožef Stefan Institute \\
  Ljubljana, Slovenia \\
  \texttt{ljupco.todorovski@fmf.uni-lj.si}
}

\begin{document}

\maketitle

\begin{abstract}
We introduce \emph{MLFMF}, a collection of data sets for benchmarking recommendation systems used to support formalization of mathematics with proof assistants. These systems help humans identify which previous entries (theorems, constructions, datatypes, and postulates) are relevant in proving a new theorem or carrying out a new construction. Each data set is derived from a library of formalized mathematics written in proof assistants Agda or Lean. The collection includes the largest Lean~4 library Mathlib, and some of the largest Agda libraries: the standard library, the library of univalent mathematics Agda-unimath, and the TypeTopology library. Each data set represents the corresponding library in two ways: as a heterogeneous network, and as a list of s-expressions representing the syntax trees of all the entries in the library. The network contains the (modular) structure of the library and the references between entries, while the s-expressions give complete and easily parsed information about every entry.
We report baseline results using standard graph and word embeddings, tree ensembles, and instance-based learning algorithms. The MLFMF data sets provide solid benchmarking support for further investigation of the numerous machine learning approaches to formalized mathematics. The methodology used to extract the networks and the s-expressions readily applies to other libraries, and is applicable to other proof assistants. With more than $250\,000$ entries in total, this is currently the largest collection of formalized mathematical knowledge in machine learnable format.
\end{abstract}

\section{Introduction}

Applications of artificial intelligence to automation of mathematics have a long history, starting from early approaches based on a collection of hand-crafted heuristics for formalizing new mathematical concepts and conjectures related to them \citep{lenat77}. In the last decade, there has been a growing interest in formalization of mathematics with \emph{proof assistants}, which verify the formal correctness of mathematical proofs and constructions, and help automate the tedious parts. The trend is correlated with the interest of machine learning community in aiding formalization efforts with its expertise.

Machine learning methods are often used to address \emph{premise selection}, i.e., recommendation of theorems that are useful for proving a given statement. DeepMath \citep{alemi16} proposes using convolutional and recurrent neural networks to predict the relevance of a premise for proving the given statement. While many other approaches \citep{polu20,welleck22} use transformers and general language models, \citet{paliwal20} have shown that taking into account the higher-order structure of logical expressions used in formalizing mathematics can greatly improve the performance of premise selection and automated proving. Indeed, many approaches use graph neural networks to learn from the higher-order structures, e.g., \citep{wang17}. More recently, graph neural networks have also been proven useful for explorative, unsupervised approaches to automated theorem proving with reinforcement learning \citep{bansal20,lample22}. Some of these approaches address alternative tasks, such as recommending or automatically selecting suitable \emph{proof tactics}, i.e., routines for performing a series of proof steps, applying a decision procedure, or for carrying out proof search.

Data sets of different origins have been used to evaluate the proposed approaches. \citet{welleck22} evaluate their approach on a selection of three hundred proofs included in the ProofWiki \citep{proofwiki} library of mathematical proofs written in a combination of natural language and \LaTeX{}.  \citet{polu20} use a standard library of the Metamath proof assistant. \citet{lample22} combine proofs from the Metamath library with proofs from the Mathlib library \citep{mathlib} of the Lean proof assistant. The latter has also been used for evaluating the approaches in \citep{han22}. \citet{wang17,paliwal20,bansal20} evaluate their models within the HOL Light proof assistant based on higher-order logic~\citep{harrison09}. The formalized proofs in standard HOL libraries have been transformed into a HOLStep data set for machine learning, where examples correspond to more than 2 million steps from $11\,400$ proofs \citep{kaliszyk17}. The training set includes proof steps in context (local hypotheses and the current statement being proved) and the library entry used in the step. Descriptions of the library entries are included in human-readable and machine-readable, tokenized versions. The data set has been recently upgraded to the interactive benchmark environment HoList for training automated proof systems with reinforcement learning~\citep{bansal19}.

We present a collection of data sets, MLFMF, based on libraries of formalized mathematics encoded in two proof assistants, Agda 
and Lean. 
It supports evaluation and benchmarking of machine learning approaches for recommendation systems in the context of formalized mathematics.

\begin{table}[!htb]
    \centering
    \begin{tabular}{l| p{10cm}}
       ID  & entry  \\
       \hline
       $\mathbb{N}$   & $\mathbb{N}:$ \newline \hphantom{oo}\texttt{zero}: $\mathbb{N}$ \newline\hphantom{oo}\texttt{suc(n)}: $\mathbb{N}\to \mathbb{N}$ \\
       \hline 
       $+$  & $0 + n = n$, for all $n\in \mathbb{N}$\newline $\texttt{suc}(m) + n = \texttt{suc}(m + n)$, for all $m, n\in \mathbb{N}$ \\ 
       \hline
       Lemma 1 (L1) & $m + 0 = m$ for all $m\in \mathbb{N}$.\newline This is proved by induction on $m$. \\  
       \hline
       Lemma 2 (L2) & $m + \texttt{suc}(n) = \texttt{suc}(m + n)$, for all $m, n\in\mathbb{N}$.\newline This is proved by induction on $m$. \\ 
       \hline
       Theorem (T) & $m + n = n + m$, for all $m, n\in \mathbb{N}$.\newline This is proved by induction on $m$. In the base case ($m = 0$), we need L1.
       In the induction step ($m = \texttt{suc}(\ell)$), we need L2. 
    \end{tabular}
    \caption{An example formalization of proof that the addition of natural numbers is commutative.}
    \label{tab:example}
\end{table}

We transform each library into a directed multi-graph whose nodes represent library entries (theorems, lemmas, axioms, and definitions), while edges represent the references between them. Consider the example in Table~\ref{tab:example}. It starts with a definition of the set of natural numbers with two simple constructors that define the first natural number 0 and constructs all the others inductively by asserting that a successor \texttt{suc(n)} of a natural number \texttt{n} is also a natural number. The definition of the addition of natural numbers follows their definition by asserting two simple rules for the left addition of 0 and the left addition of a successor. Note that the definition of $+$ references the definition of $\mathbb{N}$. Next, the first lemma establishes the rule for the right addition of zero as the first simple commutativity case. The second lemma establishes the right addition of a successor as the second case. The theorem at the end references the two lemmas to show (and prove) the commutativity of adding natural numbers.

The entries from Table~\ref{tab:example} are transformed into a multi-graph depicted in Figure~\ref{fig:example-network}. It contains five nodes, each corresponding to a table row. The multi-graph includes an edge from the node $+$ to the node $\mathbb{N}$, indicating the reference to the set of natural numbers in the definition of addition. It also contains the self-reference of $+$, since the second case of this definition is recursive. Similarly, there are four edges from the theorem node to the two lemma nodes and the two nodes defining natural numbers and addition thereof. The obtained data allows us to approach premise selection as a standard edge prediction machine learning task.

Furthermore, we transform each formalized entry into a directed acyclic graph that retains complete information about the entry, see Figure \ref{fig:example-dag}. By including the entire entry structures in the data sets, we make them suitable for further exploration of the utility of the state-of-the-art approaches to graph-based machine learning. A detailed description of the format is given in Sections \ref{sec:compute-graph} and \ref{sec:graph}.

 \begin{figure}[!htb]
     \centering
     \begin{subfigure}[b]{0.55\textwidth}
     \begin{tikzpicture}
        \Vertex[x=1, y=-4, size=0.8, label={$\mathbb{N}$}, RGB, color={128,200,128}]{nat}
        \Vertex[x=5, y=-4, size=0.8, label={T}, RGB, color={128,200,128}]{theorem}
        \Vertex[x=4, y=-2.5, size=0.8, label={L1}, RGB, color={128,200,128}]{lemma1}
        \Vertex[x=3, y=-5, size=0.8, label={L2}, RGB, color={128,200,128}]{lemma2}
        \Vertex[x=2, y=-2.5, size=0.8, label={$+$}, RGB, color={128,200,128}]{plus}
        \Edge[Direct, style=loosely dotted](theorem)(nat)
        \Edge[Direct, style=loosely dashed](theorem)(lemma1)
        \Edge[Direct, style=loosely dashed](theorem)(lemma2)
        \Edge[Direct, style=loosely dotted](lemma1)(nat)
        \Edge[Direct, style=loosely dotted](lemma2)(nat)
        \Edge[Direct, style=loosely dotted](lemma1)(plus)
        \Edge[Direct, style=loosely dotted](lemma2)(plus)
        \Edge[Direct, style=loosely dotted](theorem)(plus)
        \Edge[Direct, loopposition=150, loopsize=1.5cm, style=loosely dotted](plus)(plus)
        \Edge[Direct, style=loosely dotted](plus)(nat)
    \end{tikzpicture}
    \caption{The multi-graph representation of the example proof from Table~\ref{tab:example}.}\label{fig:example-network}
    \end{subfigure}\hfill
    \begin{subfigure}[b]{0.4\textwidth}
        \begin{tabular}{p{5cm}}
            \texttt{(:entry}\newline
                \hphantom{oo}\texttt{(:name} $\mathbb{N}$\texttt{)}\newline
                \hphantom{oo}\texttt{(:type (...))}\newline
                \hphantom{oo}\texttt{(:data}\newline
                    \hphantom{oooo}\texttt{(...)}\newline
                    \hphantom{oooo}\texttt{(:name} $\mathbb{N}$\texttt{.zero)}\newline
                    \hphantom{oooo}\texttt{(:name} $\mathbb{N}$\texttt{.suc)}\newline
                \hphantom{oo}\texttt{)}\newline
            \texttt{)}
        \end{tabular}
        \caption{An s-expression from which we obtained the DAG for the entry $\mathbb{N}$ in Table \ref{tab:example}.}\label{fig:example-dag}
    \end{subfigure}
    \caption{The two-part representation of a library. Library as a whole is represented as a network of references (a). Additionally, every entry is represented as a DAG
             which is shown here in its textual s-expression format (b). Note that some nodes of DAG were replaced by \texttt{(...)} for better readability.}
    \label{fig:example}
 \end{figure}

Our approach is general and can be applied to other proof assistants based on type theory. Moreover, even though Agda and Lean have quite different internal representations, the corresponding data sets use a common format that requires little or no knowledge about the inner workings of proof assistants. Thus our collection provides the machine learning community with easy access to a large amount of formalized mathematics in familiar formats that allow immediate application of machine learning algorithms. To our knowledge, MLFMF is the first and most extensive collection of data sets featuring more than one proof assistant and providing access to the higher-order structured representation of more than $250\,000$ mathematical formalization entries.

\section{Formalized Mathematics}

Formalized mathematics is mathematics written in a format that allows algorithmic checking of correctness of mathematical proofs and constructions. The programs that perform such checking are called \emph{proof assistants} or \emph{proof checkers}. An early proof checker was AUTOMATH \citep{bruijn70:_autom}, while today the most prominent assistants are Isabelle/HOL \citep{isabelle,hol_inter_theor_prover,harrison:_hol_light}, Coq \citep{coq}, Agda~\citep{agda} and Lean~\citep{lean}. They are all \emph{interactive}: As the user develops a piece of formalized mathematics the assistant keeps track of unfinished proof goals, displays information about the current goal, checks the input on the fly, and provides search and automation facilities.

The level of automation varies between different proof assistants. In Agda, which supports little automation, the user directly writes down proofs and constructions in abridged type-theoretic syntax that Agda checks and algorithmically elaborates to fully formal constructions. On the other end of the spectrum are Isabelle/HOL and Lean, where the user relies heavily on \emph{tactics}, which are routines that automatically perform various tasks, such as running a domain-specific decision procedure, applying a heuristic, or carrying out proof search.

The mathematical formalism most commonly used as the underpinning of a proof assistant is type theory, of which there are many variants~\citep{church40,ML75,CoquandT:coc}. 
The proof assistant processes the user input by disambiguating mathematical notations, applying tactics and other meta-level processing commands, and internally stores the resulting proofs, theorems, constructions, definitions, and types as expressions, or syntax trees, of the chosen type theory. These are typically quite verbose, so that checking their correctness is straightforward, but contain many more details than a user may wish to look at. 

Libraries of formalized mathematics comprise units, organized hierarchically with a module system or namespaces, each of which contains a number of entries: definitions of types, constructions of elements of types, statements and proofs of theorems, unproved postulates (axioms), as well as meta-level content, such as embedded documentation, definitions of tactics, hints for heuristics, and other automation mechanisms.

In the last decade the libraries of formalized mathematics have grown considerably, most recently with the rise of the popularity of the Lean proof assistant and the Mathlib library~\citep{mathlib,mathlib-preprint}, around which a mathematical community of several thousand mathematicians has formed. Such growth presents its own challenges, many of which are of the software engineering kind and can be so addressed. In our work we addressed the specific problem of \emph{recommendation}: given a large body of formalized mathematical knowledge, how can the proof assistant competently recommend theorems or constructions that are likely useful in solving the current goal? There are two typical scenarios: the user knows which theorem they would like to use but have a hard time finding it in the library, or the user is not aware of the existence of a potentially useful theorem that is already available. Both are obvious targets for machine-learning methods.

\section{MLFMF Data Sets}

In this section we describe our data sets in detail. We first explain the semantic content of the data extracted from libraries of formalized mathematics, describe the format and information content of the data sets, continue by reviewing the machine learning tasks for which the data sets were built, and finish with an overview of the technical aspects of the library-to-data-set transformation process.

\subsection{The Extracted Data}

Formalized mathematics is written by the user in a domain-specific language, often called the \emph{vernacular} or the \emph{meta-language}. The proof assistant evaluates the source code, which involves executing tactics, decision procedures, etc., verifies that the proofs and constructions so generated are mathematically valid, and stores the results using an internal type-theoretic format. One may apply machine learning techniques directly on the vernacular, as written by the user, or on the formal representation of mathematics. The former approach roughly corresponds to learning how to \emph{do} formalized mathematics, and the latter what formalized mathematics \emph{is}.

We took the latter approach, namely learning on the formalized mathematics itself, for two reasons. First, because we aimed at a uniform approach that is applicable to most popular proof assistants, it made sense to use the internal type-theoretic representations, which are much more uniform across proof assistants than the vernaculars. Second, the vernacular contains meta-level information, such as what tactics to use, from which one cannot discern directly which theorems are actually used in a given proof. Without this information, one can hardly expect a recommendation system to work well.

Every data set that we prepared is generated from a library of formalized mathematics. Most libraries, and all that we incorporated, are organized hierarchically into modules and sub-modules, each of which is a unit of vernacular code that, once evaluated by the proof assistant, results in a list of \emph{entries}: definitions of types, constructions of elements of types, theorems and their proofs, and unproved postulates. The entries refer to each other and across modules, possibly cyclically in case of mutually recursive definitions.

The internal representations of entries vary across assistants, but all have certain common features:
\begin{enumerate}
    \item Each entry has a \emph{qualified name} $M_1.M_2 \ldots M_k.N$ by which it is referred to, where $M_1.M_2\ldots M_k$ is a reference to a module in the module hierarchy and $N$ is the local name of the entry, for example $\mathtt{Algebra}.\mathtt{Group}.\mathtt{FirstIsomorphismTheorem}$.
    \item Each entry has an associated \emph{type} $T$, which specifies the information content of the body of the entry. For example, the type $\mathtt{List}(\mathbb{N})$ specifies that the entry is a list of natural numbers. Importantly, logical statements are just a special sort of types, so that the type of a proof is the logical statement that it proves. (This is to be contrasted with first-order logic, where logical statements are strictly separated from types.)
    \item An entry has a \emph{body}, which is an expression of the given entry type. In some cases the body may be missing, for instance if the user declares an axiom.
    \item Depending on the proof assistant, various \emph{meta-level information} is included, such as which arguments to functions are implicit (need not be provided by the user).
\end{enumerate}

\subsection{Data Description}

In this section, we describe the data set. We start with a brief description of data transformation process, continue with the detailed description of the resulting pair of computational graphs for the entries in the library, and the directed, multi graph of references in the library (see \ref{sec:compute-graph}) and \ref{sec:graph}).

Every data set consists of two parts. The first part is a set $\mathcal{T}$ of abstract syntax trees (AST) that correspond to the entries in the library, while the second is a directed multi-graph $G(V, E)$, where $V$ is a set of library entries, and $E$ includes the references among them. ASTs are actually trees in the case of Agda libraries. However, in Lean, they are actually directed acyclic graphs (DAGs) due to memory optimization and node-sharing: all the parents that would potentially reference their own copy of a node (or a subtree), rather reference the same node. For this reason, we refer to them as computational graphs. They provide the full information about every entry in the library and are given in the s-expression format that is much easier to parse, as compared to the typically very flexible syntax of proof assistants that allows for implicit arguments, mix-fix notation, etc. For example, the function \texttt{if\_then\_else x y z} in Agda can be called as \texttt{if x then y else z}. Learning from the source code would put an additional burden on the machine learning algorithm. Learning directly from computational graphs, on the other hand, is much easier.

\begin{figure}[!tb]
\begin{tikzpicture}
 \node[scale=3] at (3,4) (lib) {\faInstitution};
 \node[scale=1.5] at (3,3) {LIBRARY};
 \node[scale=1, text width=4.75cm,align=center, rectangle, rounded corners, draw] at (8.25, 4) (sexp)
    {
    \texttt{(name1 declaration1 body1)}\\
    \texttt{(name2 declaration2 body2)}\\
    $\vdots$\\
    };
  \node[scale=1, draw=none] at (13, 4) (graph) {};
  \path[->] (lib) edge node[above] {lib2sexp} (sexp);
  \path[->] (sexp) edge node[above] {sexp2graph} (graph);
    
  \Vertex[x=13.5, y=4.5, size=0.5, shape=rectangle]{lib}
  \Vertex[x=13.5, y=3.5, size=0.4, shape=diamond]{module}
  \Vertex[x=15, y=4.5, size=0.4, shape=circle, RGB, color={128,200,128}]{e1}
  \Vertex[x=15, y=3, size=0.4, shape=circle, RGB, color={128,200,128}]{e2}
  \Edge[Direct, style=solid](lib)(module)
  \Edge[Direct, style=dash dot dot](module)(e1)
  \Edge[Direct, style=dash dot dot](module)(e2)
  \Edge[Direct, style=loosely dotted](e1)(e2)
\end{tikzpicture}
\caption{The two stages of the data transformation. First, a language-dependent (i.e., Agda or Lean) command line tool is used to transform the library entries into s-expressions. In the second stage, Python scripts are used to explicitly construct the directed multi-graph, which contains library modules, entries, and references among them.}
\end{figure}
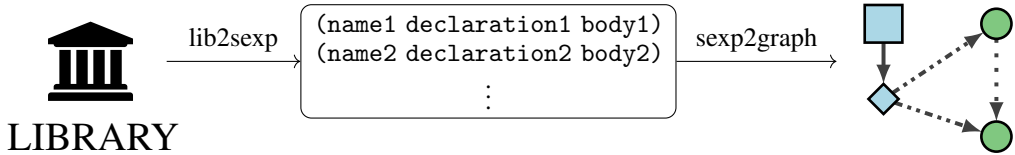

\begin{figure}[h!]
    \centering
    \begin{tikzpicture}
        \Vertex[x=0, size=0.8, label={Entry}, RGB, color={128,200,128}]{e}
        \Vertex[x=-2, y=-0.8, size=0.8, label={name}, RGB, color={75,175,128}]{n}
        \Vertex[x=0, y=-2, size=1.2, label={declaration}, RGB, color={75,175,128}, style={isosceles triangle, minimum width=1cm, rotate=90, minimum height=2cm}]{d}
        \Vertex[x=2.25, y=-0.6, style={draw=none}, RGB, color={255, 255, 255}]{fake}
        \Vertex[x=2, y=-2, size=1, label={body}, RGB, color={75,175,128}, style={isosceles triangle, minimum width=1cm, rotate=90, minimum height=2cm}]{b}
        \Edge[](e)(n)
        \Edge[](e)(d)
        \Edge[](e)(fake)
    \end{tikzpicture}
    \caption{The DAG representing an entry has a single root node with three children: a node containing the entry name, a DAG containing the entry declaration, and a DAG representing the entry body.}
    \label{fig:ast}
\end{figure}
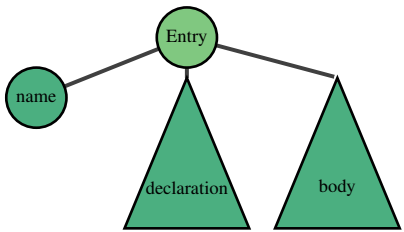

\subsection{The Computational Graphs}\label{sec:compute-graph}
During compile time, the full type of every entry in the library is computed and the source code of the entry is converted to a (directed acyclic) computational graph. We intercept this procedure and export every entry as a Lisp \textit{s-expression}, which is defined recursively as:
\begin{enumerate}
    \item A literal is an s-expression, and
    \item A list of s-expressions is an s-expression.
\end{enumerate}
For example, the literals \texttt{12} and \texttt{"foo"} are s-expressions, and the list \texttt{("foo" ("bar" 12) "baz")} is also an s-expression with three elements: \texttt{"foo"}, \texttt{("bar" 12)} (which contains two s-expressions) and \texttt{"baz"}. Every s-expression that is obtained from the entries in a library is three-part, as shown in Fig.~\ref{fig:ast}. In consists of the name of the entry, the s-expression that describes the declaration, and the s-expression that describes the body of the entry.

Even though the entries were manually encoded and mostly take at most a few kilobytes of space, their computational graphs can be much larger (more than a gigabyte), mostly due to the type checking and the expansion of the declared type of the entry.


\subsection{The Multi-Graph}\label{sec:graph}
\begin{figure}[b!]
     \centering
     \begin{subfigure}[b]{0.99\textwidth}
        \centering
         \begin{tikzpicture}
            \Vertex[x=-3, size=1.5, label={Library}, shape=rectangle]{lib}
            \Vertex[x=2, size=1.5, label={Module}, shape=diamond]{m}
            \Vertex[x=7, size=1, label={Entry}, RGB, color={128,200,128}]{e}
            \Edge[Direct, label={CONTAINS}, style={solid}](lib)(m)
            \Edge[Direct, label={CONTAINS}, style={solid}, loopposition=90, loopsize=1.2cm](m)(m)
            \Edge[Direct, label={DEFINES}, style={dash dot dot}](m)(e)
            \Edge[Direct, label={REFERENCE FROM DECLARATION}, loopposition=90, loopsize=2cm, style={text width=1.5cm, align=center, loosely dotted}](e)(e)
            \Edge[Direct, label={REFERENCE FROM BODY}, loopposition=270, loopsize=2cm,style={text width=1.5cm, align=center, loosely dashed}](e)(e)
        \end{tikzpicture}
        \caption{The meta-structure of the graph.}
        \label{fig:meta-graph}
    \end{subfigure}
     \hfill
     \begin{subfigure}[b]{0.99\textwidth}
         \centering
         \begin{tikzpicture}
            \Vertex[x=-3, y=-4, size=1.5, label={stlib}, shape=rectangle]{stdlib}
            \Vertex[x=0, y=-4, size=1.5, label={Function}, shape=diamond]{func}
            \Vertex[x=3, y=-3, size=1.5, label={Bijection}, shape=diamond]{bijMod}
            \Vertex[x=8, y=-3, size=1, label={Bijection}, RGB, color={128,200,128}]{bijEntry}
            \Vertex[x=5, y=-4, size=1, label={id}, RGB, color={128,200,128}]{id}
            \Vertex[x=3, y=-5, size=1.5, label={Injection}, shape=diamond]{injMod}
            \Vertex[x=8, y=-5, size=1, label={injective}, RGB, color={128,200,128}]{injEntry}
            \Edge[Direct, style=solid](stdlib)(func)
            \Edge[Direct, style=solid](func)(bijMod)
            \Edge[Direct, style=solid](func)(injMod)
            \Edge[Direct, style=dash dot dot](bijMod)(bijEntry)
            \Edge[Direct, style=dash dot dot](bijMod)(id)
            \Edge[Direct, style=dash dot dot](injMod)(injEntry)
            \Edge[Direct, style=loosely dotted](id)(bijEntry)
            \Edge[Direct, style=loosely dashed](id)(injEntry)
        \end{tikzpicture}
        \caption{An excerpt from Agda's standard library. The entry \texttt{id} (indentity function) is defined in the module \texttt{Bijection}, which is a submodule of the module \texttt{Function}, which is one of the top modules of the library. In the declaration of \texttt{id}, the entry \texttt{Bijection} is referenced. In the body of \texttt{id}, the entry \texttt{injective} is referenced. The two referenced entries are defined in the submodules \texttt{Bijection} and \texttt{Injection}, respectively.}
    \label{fig:stdlib-excerpt}
     \end{subfigure}
    \caption{A meta-graph of libraries (a) and a subgraph of the graph that was created from Agda's standard library (b) that follows the prescribed meta-structure.}
    \label{fig:library}
\end{figure}
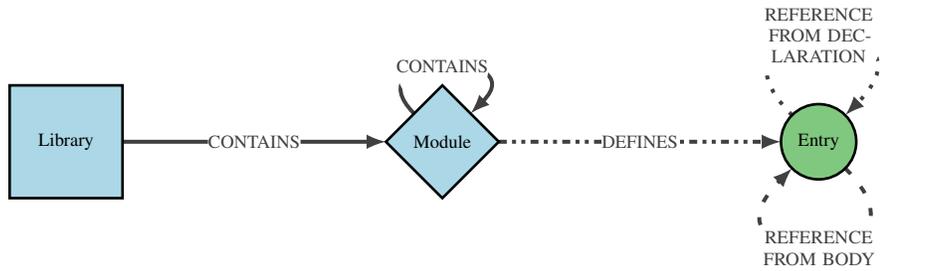

For simplicity reasons, we will refer to the directed multi-graph $G(V, E)$ simply as a graph. Its meta-structure is shown in Fig.~\ref{fig:meta-graph}. In the description below, we follow this structure (in the bottom-up manner) and the concrete example of a subgraph for Agda's standard library in Fig.~\ref{fig:stdlib-excerpt}.

\paragraph{Entry nodes.} Every module in a library defines at least one entry (shown as green circles), e.g., \texttt{Bijection DEFINES id}, \texttt{Bijection DEFINES Bijection} (these are two different nodes), and \texttt{Injection DEFINES injective}. We further differentiate between different kinds of entries, as shown in Tab.~\ref{tab:entry-kinds}.
\begin{table}[!htb]
    \centering
    \caption{Tags of the nodes in s-expressions.}
    \label{tab:entry-kinds}
    \begin{tabular}{l | p{10cm}}
			kind & description \\
			\hline
			\texttt{:data} & inductive data type (natural numbers, lists and trees)\\
			\texttt{:constructor} & data-type constructor (successor, cons)\\
			\texttt{:function} & function (including constants as nullary functions) \\
			\texttt{:record} & record type (a structure with named fields or attributes)\\
			\texttt{:axiom} & postulated type or statement (no inhabitant or proof given) \\
			\texttt{:primitive} & built-in (primitive) function \\
			\texttt{:sort} & the sort of a type (proposition, universe at a given level) \\
            \texttt{:recursor} & the induction/recursion principle associated with an inductive data-type \\
            \texttt{:abstract} & entry whose body is hidden
		\end{tabular}
\end{table}
Most of the nodes in the graph are entries (and most of them are \texttt{function}s), and most of the edges are of type \texttt{REFERENCE FROM DECLARATION/BODY}). 

\paragraph{Library and module nodes.} The only nodes with no incoming edges (root nodes) are the library nodes (shown as blue squares). Every graph contains at least one library node---the one that corresponds to the library itself. In Fig.~\ref{fig:stdlib-excerpt}, this is the node \texttt{stdlib}. However, the graph might contain an additional node \texttt{outer library} if any of the library entries reference some external entries that are not part of the library (for example, built-in types). The library nodes are directly connected to the nodes representing modules (shown as blue diamonds) via the edges of the type \texttt{CONTAINS}, e.g., \texttt{stdlib CONTAINS Function}. Every module can contain zero or more (sub)modules, e.g., \texttt{Function CONTAINS Bijection}.

In the case of Agda libraries, the module nodes correspond to the modules that are actually present in the library and resemble the file system of the library. Lean, however, supports the use of namespaces. If the file \texttt{a/b/c.lean} defines an entry \texttt{foo.bar.F}, and the file \texttt{d/e.lean} defines an entry \texttt{foo.bar.G}, those two entries are part of the same namespace \texttt{foo.bar} and the exact location in the file system where these two entries were defined, is irrelevant. Therefore, module nodes for Lean's library Mathlib4 correspond to namespaces in the library. Following the previous example, we create module nodes \texttt{foo} and \texttt{bar}, together with the edge \texttt{foo CONTAINS bar}.

\subsection{Machine Learning Tasks}
The main motivation for the creation of the data set was the development of machine learning algorithms that would enhance current proof assistants and help mathematicians using them. This translates to the following two machine learning tasks.

\paragraph{Link prediction.} Given the current state of the multi-graph of references among the entries, learn a model that predicts the future, novel links (references) among the library entries. Formally, we learn a model $M: (u, v)\mapsto M(u, v) \in [0, 1]$ that given two nodes $u$ and $v$ outputs the model confidence in the presence of the edge $(u, v)$. The (current) computational graphs of the entries can be used as additional information for learning such a model. If learning from the multi-graph only, one can use standard node- or edge-embedding approaches as well as graph neural networks.

\paragraph{Recommendation.} The problem of predicting the future references among the entries could be understood as a recommendation task as well. Given a specific unfinished entry (possibly with some additional context, such as the list of lemmas/ claims that were used last), the task is to recommend the candidates that could be referenced in the current computational graph of the entry to complete it.

Note that the two tasks are equivalent, i.e., solving one solves the other. A link prediction model $M$ (see above) can be converted into a recommendation system by fixing the entry $u$ and recommending the entries $v \in V$ with the highest confidence levels $M(u, v)$. Vice versa, given a recommendation model $M': u \mapsto M'(u)\subseteq V$, we can define a corresponding link-prediction model $M$ as $M(u, v) = 1$ if $v\in M'(u)$, and $M(u, v) = 0$ otherwise.

Since the essential part of the MLFMF data set is a directed multi-graph (which represents a heterogeneous network), other standard learning tasks for graphs/ networks might also be interesting. Here, we mention two example instances of the common node classification task.

\paragraph{Entry class detection.} A straight-forward instance of node classification task would be classifying the entries into their types from Table~\ref{tab:entry-kinds}, e.g., \texttt{function} or \texttt{axiom}. This does not require additional manual labeling and should not be too hard, especially when computational graphs are taken into account, since the structure of a \texttt{function} is quite different from the structure of, e.g., \texttt{record}.

\paragraph{Claim detection.} A more challenging instance of node classification is predicting whether a \texttt{function} entry is a claim (e.g., a lemma, corollary, theorem, etc.) or not, since some of the entries are simply definitions of, for example, the addition of natural numbers. Approaching this task, however, would require additional (manual) labeling of the entries.

\subsection{License}
We make MLFMF publicly available under the Creative Commons Attribution 4.0 International\footnote{https://creativecommons.org/licenses/by/4.0/} (CC BY 4.0) license at \url{https://github.com/ul-fmf/mlfmf-data}.

\section{Experiments and Results}

In this section, we first introduce the experimental setup for the baseline experiments (how to prepare the train and test part of the data set, and which standard metrics can be used), and then, after briefly introducing the baseline methods, we report the experimental results.

\subsection{Train-test split}
When splitting the graph into train and test data sets, we should split the multi-graph $G(V, E)$ and the set of computational graphs of the entries. In the case of the link prediction and recommendation tasks, we should focus on \texttt{function} nodes, since these are the only nodes that correspond to a computational graph whose body contains a proof of a claim formalized in the declaration part of the computational graph.

In our baseline experiments from Sec.~\ref{sec:results}, we follow a generic approach to creating a train-test split. The approach takes two parameters: $p_\text{test} \in (0, 1)$, $p_\text{body} \in [0, 1)$. First, we randomly choose the proportion $p_\text{test}$ of \texttt{function} nodes. We assume that those correspond to partially written entries whose computational graphs have completely specified type, i.e., the user knew how to formalize a claim, and \emph{partially} known body, i.e., the proof of the claim is not finished yet. Note that, often, proofs are not written linearly and might contain so-called holes at problematic parts where the right lemmas are yet to be applied (possibly with already known arguments). Thus, we need to modify the computational graphs of the test nodes to reflect the changes in the multi-graph.

We simulate the applications of the missing lemmas by keeping only the proportion of $p_\text{body}$ of the references in the body. Since our graph contains a weighted edge \texttt{u REFERENCE FROM BODY v}, which we either remove or keep intact, we remove all references to $v$ from the body of $u$ or none of them. Then, the unfinished proofs are simulated by keeping the proportion of $p_\text{body}$ of the body of $u$, which is done by iterative pruning of the leaves of the body. At each iteration, a leaf is chosen uniformly at random. If the chosen leaf is a reference that we have to keep, the leaf is not pruned and we continue with the next iteration.

The removed edges represent positive test examples, and the negative test examples for learning predictive models need to be sampled. In the baseline experiments, the negative test examples were sampled uniformly at random.

\subsection{Evaluation metrics}
For link prediction, one can use standard classification metrics, such as accuracy, precision, recall, and $F_1$-score. If the model returns its confidence $M(u, v)\in [0, 1]$ instead of the class value ($M(u, v)\in \{0, 1\}$), one could additionally consider area under the receiver-operating-characteristic curve. Similarly goes for the recommendation models: one can use precision and recall.

If the recommendation model returns the relevance score of a candidate entry to the current context, we can rank candidates according to the score values, with the top recommendation having a rank of one (1). We can then compute the minimal (and the mean) rank of the actual references and average them over the testing examples. This is an important metric, since it counts the number of false recommendations with better ranks than any of the actual references. Ideally, the minimal rank is close to one, i.e, the top-ranked recommendation mostly matches the missing entry to be referenced.

\subsection{Baseline Methods}


\paragraph{Dummy recommender.} This recommender ignores the current context and always recommends the $k$ nodes of the multi-graph with the highest in-degree.

\paragraph{Bag of Words/TFIDF recommenders.} Bag of Words (BoW) recommender converts every computational graph $g(u)$ of an entry $u$ in a library into a bag of words $\operatorname{BoW}(u)$. We compute the relevance $M(u, v)$ of the candidate entry $v$ for the current context $u$ using the Jaccard similarity between the corresponding bag-of-words:
    $$
    J(\operatorname{BoW}(u), \operatorname{BoW}(v)) = \frac{| \operatorname{BoW}(u) \cap \operatorname{BoW}(v) |}{| \operatorname{BoW}(u) \cup \operatorname{BoW}(v) |}.
    $$
Similarly, TFIDF-recommender embeds $g(u)$ into a term-frequency-inverse-document-frequency vectors (obtained from the corresponding bags-of-words) as implemented in Scikit-Learn 1.2.2 \citep{scikit-learn}. The relevance of the candidate entry $v$ is computed as a Manhattan or a cosine distance between the TFIDF-vectors of $u$ and $v$.

\paragraph{FastText embedding recommender.} It embeds every computational graph $g(u)$ into a vector $\vec{\varphi}(g(u)) =
        \sum_{\text{word}\in g(u)}
            w(\text{word}, g(u)) \cdot \varphi_\text{cc}(\text{word})$,
where $\varphi_\text{cc}(\text{word})$ is the vector of \texttt{word} obtained from the fastText model trained of Common Crawl \citep{fastText2018}, and $w(\text{word}, g(u))$ is the TFIDF weight of the word in the entry $u$.

\paragraph{Recommendations via analogies.} We design a recommender that is based on fastText analogies property, i.e., the fact that $x = \text{queen}$ is one of the approximate solutions of
    $\varphi_\text{cc}(\text{king}) - \varphi_\text{cc}(x) = \varphi_\text{cc}(\text{man}) - \varphi_\text{cc}(\text{woman})$.
We design the \emph{analogy recommender} that for a given entry $u$ recommends the nodes $v$, for which a good analogy $u'\to v'$ of the edge $u\to v$ can be found. The relevance of the candidate entry $v$ in a given context $u$ is defined in terms of the Manhattan distance as
$$
    r(u, v) = 1\quad / \min_{u'\to v'\in E(G)} \| [\varphi_\text{cc}(u) - \varphi_\text{cc}(v)] - [\varphi_\text{cc}(u') -  \varphi_\text{cc}(v')] \|_1.
$$

\paragraph{Node2vec-based link prediction.} We train a node2vec \citep{node2vec} model (as implemented in Gensim 4.3.1 \citep{rehurek2011gensim} on the multi-graph to obtain node embeddings. We obtain the embedding of the edge $(u, v)$ by concatenation of the node embeddings for $u$ and $v$. A tree-bagging classifier $M: \varphi(u\to v)\mapsto M(\varphi(u\to v))\in [0, 1]$ is trained on the tabular data obtained with using the edge embeddings as inputs and the edge presence as the target to be predicted. We selected bag of trees ensemble since it is a robust classifier, working well on tabular data, when using the recommended settings of 100 fully grown (not pruned) classification trees.

We selected baseline methods that are not computationally expensive and are robust to hyper-parameter settings: if not mentioned otherwise, the methods use the default parameter settings. We can combine multiple embeddings (e.g., those from node2vec together with those from fastText) as the input to the similarity measure of the recommender or classifier for the link prediction task. However, as noted in the next section, this did not improve the best results. In all the experiments, we generated the train-test split with the parameters $p_\text{test} = 0.2$ and $p_\text{body} = 0.1$. The results here are reported for $k = 5$ recommended items and the threshold $\vartheta = 0.5$ for classification.

\subsection{Results}\label{sec:results}

The experiments on Lean were run on a computer with 4 Intel Core i7-6700K CPU cores and 64 GB of RAM. The experiments on Agda were run on a smaller machine (2 Intel Core i7-5600U CPU cores, 12~GB of RAM). Experiments that would last more than a week were not carried out (analogies on the Type Topology and Mathlib4 libraries, and fastText on the Mathlib4 library).

Tab.~\ref{tab:ressults} reports the results of the experiments: for extended report including other evaluation metrics (accuracy@k, area under the ROC curve, etc.), and ablation study of node2vec on Agda stldib, check the supplementary material. For the algorithms that were run with more than one parameter setting, the best results are reported (for example, TFIDF was run with cosine- and Manhattan-based similarity measures). The best-performing baseline method is node2vec. It is the only one that ranks on average at least one actual reference among the ten most relevant candidate references for the three Agda libraries. However, it fails to do so for Lean Mathlib4 and this can be only partially explained by the size of the Mathlib4. Note that node2vec is also the only one that explicitly learns from the multi-graph and, apparently, humans writing proofs in Agda, structured the references better than the computers in Lean, where built-in search heuristics (tactics) are used. The multi-graph is partially used by the analogies recommender as well, since the candidate recommendations are evaluated by considering the existing references $u'\to v'$ in the library. This might be the reason for its good performance on the Agda stdlib data set.

\begin{table}[htbp]
  \caption{The accuracy (acc) and minimal rank of the true reference for the MLFMF data sets. The best results (bold) are obtained with a combination of a node2vec and a tree-bagging classifier.}
  \smallskip
  \centering
    \begin{tabular}{l|rr|rr|rr|rr}
    \multicolumn{1}{c|}{\multirow{2}[4]{*}{method}} & \multicolumn{2}{c|}{Agda stdlib} & \multicolumn{2}{c|}{Agda unimath} & \multicolumn{2}{c|}{Agda TypeTopology}  & \multicolumn{2}{c}{Lean Mathlib4} \\
    & \multicolumn{1}{c}{acc} & \multicolumn{1}{c|}{minRank} & \multicolumn{1}{c}{acc} & \multicolumn{1}{c|}{minRank} & \multicolumn{1}{c}{acc} & \multicolumn{1}{r|}{minRank} & \multicolumn{1}{c}{acc} & \multicolumn{1}{c}{minRank} \\
    \hline
    Dummy & 0.51  & 218   & 0.53 &  2134  & 0.50  & 4556    & 0.51    & 26065  \\
    BoW   & 0.50      & 1608  & 0.50  & 1571  & 0.50  &  4496     &   0.50    & 15458  \\
    TFIDF & 0.51    & 144   & 0.52  & 112   & 0.51  & 552   &   0.51    & 443  \\
    fastText &  0.51   & 132   & 0.52  & 394   &  0.50    & 1292      &  NA     & NA  \\
    analogies & 0.52      & 37    & 0.51  & 158   &  NA     &  NA     &   NA    & NA  \\
    node2vec & \textbf{0.96} & \textbf{4.37} & \textbf{0.96} & \textbf{3.24} & \textbf{0.98} & \textbf{5.81}  &   \textbf{0.95}    & \textbf{195} \\
    \end{tabular}%
  \label{tab:ressults}%
\end{table}%

Surprisingly, TFIDF embeddings perform no worse (or even better) than FastText embeddings. The reason for this might be that many words, such as group, ring, etc.~have different meanings in mathematics than in general texts. Note that we tried to run additional experiments with the combination of node2vec and TFIDF/fastText embeddings, but accuracy and minRank were both worse, as compared to the node2vec results.

In sum, the baseline results show that the information on the structure of the multi-graph is crucial for obtaining classifiers with performance beyond the default performance of the dummy baseline. The recommendation performance, measured as mean minimal rank, is valuable enough (less than five recommendations to be checked to find the right one) for two Agda libraries. Developing sound recommendation systems for the other two libraries remains a challenge to be addressed by machine learning methods beyond the baselines considered here.

\section{Conclusion}

We introduced MLFMF, a suite of four data sets corresponding to three libraries in Agda and one library in Lean proof assistants. It includes almost $250\,000$ entries, i.e., definitions, axioms, and theorems with accompanying proofs. References between entries are included in a multi-graph, where nodes are entries, edges represent references among the entries, and each entry is represented with a direct acyclic graph reflecting the structure of the entry source code encoding. Such a structure provides machine learning researchers with an opportunity to address the task of recommending relevant entries for the goal at hand as a standard edge prediction task. Such a representation of the entries allows for use of graph-based methods that can exploit the structural and semantic information stored in the multi-graphs. The report on the results of the baseline methods establishes a benchmark for comparative evaluation of future developments of machine learning for mathematical formalization that goes beyond a single proof assistant.

A notable limitation of our data sets is the lack of information on the developmental evaluation of the libraries. If the latter could be followed, the more realistic test nodes $V_\text{test}$ could be defined as the \emph{latest} $|V_\text{test}|$ nodes encoded in the library. However, even with version control information (available since the libraries are stored in GitHub repositories but currently not included), determining the entries’ chronological order might be computationally expensive. An approximation of the chronological order might be obtained by computing a topological ordering on the \texttt{function} nodes and selecting nodes from the tail of the ordered list. However, existing definitions in a library might be rewritten, so the accuracy of such an approximation is questionable.

Finally, we plan to include other, more recent libraries in our data set collection. The newly incorporated libraries might include references to earlier, standard libraries, providing further opportunities for real-world testing scenarios.

\section{Acknowledgments}

This material is based upon work supported by the Air Force Office of Scientific Research under award number FA9550-21-1-0024. The authors also acknowledge the financial support of the Slovenian Research Agency via the research core funding No.~P2-0103 and No.~P1-0294.

\bibliographystyle{plainnat}
\bibliography{references}

\end{document}


\maketitle
This document provides several pieces of meta-information about the MLFMF data set collection, as well as some additional details and results from the experiments.

All the code is available at the GitHub repository \url{https://github.com/ul-fmf/mlfmf-data}. For a detailed description of the preprocessing scripts and the script for running the model, please refer to the README in the repository. However, due to space limitations, all the preprocessed data can be found in the anonymous Google Drive folder \url{https://drive.google.com/drive/folders/1o6QP3Vo_9KR5gB8GPYKeSXDW8pPN8i2U?usp=sharing}, which -- for self-sufficiency reasons -- contains the relevant preprocessing tools and a simple script to load the data sets.

\section{Data Collection}
We obtained the source code of the libraries from their publicly available GitHub repositories. At the time of collection, we retrieved the latest versions of the libraries, which are specified in Table~\ref{tab:data-version}.
In the case of Agda, a fork (\url{https://github.com/andrejbauer/agda/tree/master-sexp}) of the official Agda repository (\url{https://github.com/agda/agda}) was created to modify the compilation procedure, so that it outputs s-expressions (see the main article, Section~3). In the case of Lean, a standalone tool (\url{https://github.com/andrejbauer/lean2sexp}) was developed and implemented in Lean to process entries in Lean's \texttt{olean} binary format.

\begin{table}[htbp]
  \centering
  \caption{The versions of the four libraries transformed to the data sets.}
    \begin{tabular}{l|l}
    GitHub repository & commit\\
    \hline
     agda/agda-stdlib & cfa2504316d7e03e4a1a5f2353976796e07a9f1e \\
     UniMath/agda-unimath & bca22ae30b07f9ee02f82d1f4893ddc389185cb0 \\
     martinescardo/TypeTopology & 8b44abc0c99775c8141709a47a52ec5827357886 \\
    leanprover-community/mathlib4 & cdbd878af82f017d6a1db38e5581a348d7706002 \\
    \end{tabular}%
  \label{tab:data-version}%
\end{table}%

A co-author of this paper developed both conversion procedures, and their anonymized implementations (without the links to the specific GitHub repositories) are present in the Google Drive folder mentioned above. We plan to make them publicly available upon acceptance and/ or public release of the data set.

\section{Data Description}
As described in the article manuscript (see Section~3.2), every data set corresponds to a library of formalized mathematics. The data set consists of two parts: a set of computational graphs of the library entries (described in Section~3.3 of the article) and a directed multi-graph (see Section~3.4). We store every computational graph in a separate \texttt{.dag} file (a text file whose extension \texttt{dag} stands for directed acyclic graph). The second part of the data set, the directed multi-graph, or network for short, is stored in a text file, listing its nodes and links.

In the following subsections, we describe the file format and the structure of the repository.

\subsection{The Structure of the DAG Files}
Every DAG file contains a tab-separated table with four columns of \texttt{NODE ID}, \texttt{NODE TYPE}, \texttt{NODE DESCRIPTION} and \texttt{CHILDREN IDS}, described in Table~\ref{tab:dag-columns}. Excluding the header, every line in the file represents a single DAG node. For example, the row 
\begin{center}
    \texttt{2923823	:name	"Relation.Binary.Definitions.Transitive 70"	[]}
\end{center}
describes a node with the ID \texttt{2923823}, which is a\texttt{:name} that refers to the fully qualified name\footnote{In Agda there might be name duplicates, so an additional (Agda-provided) id of the entry was included into the name (\texttt{70} in the example above).} \texttt{"Relation.Binary.Definitions.Transitive 70"}, and has no children (empty list \texttt{[]}).

\begin{table}[!htb]
    \caption{The meaning of the four columns in the \texttt{.dag} files.\label{tab:dag-columns}}
    \begin{center}
        \begin{tabular}{l|p{10cm}}
            column name & description \\
            \hline
            \texttt{NODE ID} &
            A unique identifier (integer) of a node in the corresponding computational graph. The uniqueness is guaranteed (and meaningful) locally within a single DAG file. The same \texttt{NODE~ID} in different files does not denote that the corresponding nodes are the same or related.\\
            \rule{0pt}{3ex}
            \texttt{NODE TYPE} &
            A type of the node revealing its specific role in the computational graph, e.g., \texttt{:entry} declares a start of the new entry, \texttt{:name} specifies the name of the entry or its reference.\\
            \rule{0pt}{3ex}
            \texttt{NODE DESCRIPTION} &
            Additional information about the node. This might be the name that a \texttt{:name} node introduces or the value of the literal in the \texttt{:literal} nodes. For most of the node types, it is empty.\\
            \rule{0pt}{3ex}
            \texttt{CHILDREN IDS} & A list of unique identifiers of the children of the current node in the computational graph.
        \end{tabular}
    \end{center}
\end{table}

Even though the data were prepared uniformly (as much as possible) for both proof assistants, Agda and Lean, only 20 node types are relevant for both Agda and Lean. Additional 45 are Agda-specific node types, while 19 are Lean-specific. Note that we included the node type for completeness and lossless transfer of the information available in the libraries. For many machine learning applications, the information on the node type can be ignored. In our experiments with the baseline methods (Section 4.3 of the article), we did not use the information on the node type. However, machine learning experts familiar with the detailed semantics of the definition types in the programming languages Agda and Lean can use the information about the node types.

\subsection{The Structure of the Network Files}
The network file lists the nodes and the links of the multi-graph of references among the entries and modules of the corresponding library. Lines starting with the word \texttt{node} represent nodes with two tab-separated fields
\begin{center}
    \texttt{node	<node name>	<node properties>}.
\end{center}
The entry names are unique and include the modules (Agda) and namespaces (Lean) in the context in which they are defined. The only property of a node is its \texttt{label}, whose value is either one of the nine entry types given in Table~1 of the article, a \texttt{:module} or a \texttt{:library}. The latter two denote nodes corresponding to the library modules (in Agda) and the current library or the referenced libraries.

The lines starting with the word \texttt{link} are of form
\begin{center}
    \texttt{link	<source node>	<sink node>	<link type>	<link properties>}.
\end{center}
The first two fields refer to the source’s and sink’s node names. The following field is the link type, i.e., one of the types from Table~\ref{tab:link-types}, which also explains the link properties (the last field in the line). In addition to these link types, there are four Agda-specific link types: \texttt{REFERENCE\_<TYPE/BODY>\_TO\_<WITH/REWRITE>}. These encode the references from actual user-defined entries
(either from their types or bodies) to the entries created by the Agda compiler. For example, the definition
\begin{verbatim}
    isEven : Nat -> Bool
    isEven n with (mod2 n)
    ...           |  0     = true
    ...           |  1     = false
\end{verbatim}
(which reads as \textit{return true if $n\text{ mod } 2 = 0$, and false if $n\text{ mod } 2 = 1$}) contains a \texttt{with} block. This block is internally represented as a separate entry, referenced from the body of the entry \texttt{isEven}. These reference links have the same properties as the standard reference links from Table~\ref{tab:link-types}.

\begin{table}[t]
    \caption{The four types of the links in the network.\label{tab:link-types}}
    \begin{center}
        \begin{tabular}{l|l|p{8cm}}
            link type & property  & description  \\
            \hline
            \texttt{DEFINES} & none &
            A link from module to entry nodes denoting that the corresponding module defines the corresponding entry.\\
            \rule{0pt}{3ex}
            \texttt{CONTAINS} & none &
            A link from a library/ module node that contains another (sub-)module node.\\
            \rule{0pt}{3ex}
            \texttt{REFERENCE\_TYPE} & \texttt{w} &
            A link $a\to b$ between two entry nodes $a$ and $b$ denoting that the entry corresponding to $a$ references the entry $b$ from the type (declaration) part of its computational graph $g(a)$. The property \texttt{w} is the count of the references from the type part of $g(a)$ to entry $b$.\\
            \rule{0pt}{3ex}
            \texttt{REFERENCE\_BODY} &\texttt{w} &
            An analogue of \texttt{REFERENCE\_TYPE} denoting a reference from the body part of the computational graph $g(a)$ to the entry $b$.
        \end{tabular}
    \end{center}
\end{table}

\paragraph{External entries.} A given entry in a library might reference some built-in method of a proof assistant (as a programming language), e.g., a method on lists in Lean. In that case, the s-expression of the referenced method might not be available, but its fully qualified name is. We include the corresponding node and the corresponding external modules in the network. The only difference from the standard case is that we set the \texttt{NODE TYPE} of such modules as \texttt{:external-module}.

\subsection{Data Size}
Table~\ref{tab:data-size} gives the basic size-related statistics of the datasets. We can see that Lean Mathlib4 is approximately ten times bigger than any of Agda's libraries.
Next, most of the nodes in the network are entry nodes. The difference between the number of nodes and entries is due to the module nodes and the references to the nodes not being part of the library we processed. This often happens in Mathlib4 and sometimes in \texttt{stdlib}. The other two libraries (Unimath and TypeTopology) are---in that sense---self-contained.

\begin{table}[htbp]
  \centering
  \caption{The number of entries, the total and maximal size of their compute graphs (nodes that they contain) and the size of the network $G(V, E)$.}
    \begin{tabular}{l|rrr|rr}
    library & \multicolumn{1}{l}{entries} & \multicolumn{1}{l}{total entry size} & \multicolumn{1}{l|}{max entry size} & \multicolumn{1}{c}{$|V|$} & \multicolumn{1}{c}{$|E|$}\\
    \hline
    Agda stdlib & 16,483 & 1.3$\cdot 10^7$ & 8.1$\cdot 10^3$ & 16,855 & 242,484 \\
    Agda Unimath & 20,163 & 1.9$\cdot 10^7$ & 9.8$\cdot 10^5$ & 21,493 & 322,446 \\
    Agda Type Topology & 31,232 & 3.6$\cdot 10^7$ & 3.3$\cdot 10^5$ & 31,701 & 726,710 \\
    Lean Mathlib4 & 202,769 & 5.0$\cdot 10^8$ & 8.3$\cdot 10^6$ & 215,229 & 7,378,824 \\
    \end{tabular}%
  \label{tab:data-size}%
\end{table}%

\subsection{Additional details on the experiments}

Here, we first give some additional details on the experiments in which node2vec was used. Then, we show an extension of the Table 3 with additional quality measures.

\subsubsection{The detailed node2vec experimental setup}

\paragraph{Network preprocessing details.} Our network is a weighted directed multi-graph,
while node2vec was designed for simple (possibly directed) graphs. We carry out the conversion to the undirected weighted graph in the following steps.
\begin{enumerate}
    \item We transform the weights $w = w(u, v)$ on the directed edges $(u, v)$ via the TFIDF-like transformation.
          First, we create a document for every node $u$. The ID of every node $v$, such that $(u, v)$
          is a directed edge, appears (as a single \textit{word}) in the document $w(u, v)$ times (note that the weights are the counts of references).
          Second, the updated weight $w'(u, v)$ is the TFIDF-score of the (ID of) $v$ in the document that corresponds to the node $u$.
    \item There can be more than one directed edge from $u$ to $v$. The weights on such edges from the previous step are merged into a single value by summation.
    \item To obtain an undirected graph, the final weight on the (undirected) edge between $u$ and $v$ is the sum of the weights on the directed edges
    $(u, v)$ and $(v, u)$ (if the edges exist).
\end{enumerate}
This transformation prevents walks from visiting \emph{hubs} (nodes with extremely large degrees) too often
and, at the same time, still takes the weights into account.

\paragraph{Node2vec implementation details.}
For Agda libraries, we were able to use the freely available implementation of node2vec (\url{https://github.com/eliorc/node2vec}),
which precomputes the transition probabilities, generates the walks, and then uses Gensim's word2vec model.
However, computing the transition probabilities for a graph $G(V, E)$ takes $\mathcal{O}(|E|^2)$ space,
which was infeasible in the case of Lean Mathlib4. Therefore, we used our own implementation of walk generation
(written in Python and compiled \emph{just it time} with numba).

\subsubsection{Extended results}
Here, we first extend Table 3 from the main text (the main table with results) with additional evaluation measures.
In addition to that, we show the results of an ablation study of node2vec on Agda standard library.

\paragraph{Additional evaluation measures.} The additional performance measures are
accuracy@k where $k = 5$ (Table \ref{tab:acc-k}), precision (Table \ref{tab:precision}), recall (Table \ref{tab:recall}), and area under the ROC curve (Table \ref{tab:aucroc}). Accuracy@k is a recommender system evaluation measure that gives us the average proportion of the correct recommendations in the top-$k$ recommendations. The rest of the measures are well-known evaluation measures for classification models. AU-ROC is threshold independent, whereas the precision and recall are reported at threshold $\vartheta = 0.5$.

Node2vec achieves the best accuracy@k for the three Agda libraries, while on the Lean Mathlib4 library, the dummy classifier (still) performs better.
A similar situation can be observed with precision values, however, most of them should be considered trivial since the corresponding recalls (and the AU-ROC values as well) are rather low.

It turns out that the models are for most of the test edges $(u, v)$ quite sure, whether this edge should be present or absent, even though they are far from being perfect. For example, AU-ROC of node2vec is 0.99 on Mathlib4, but its accuracy@k is (approximately) 0.00. This means that randomly sampling negative edges did not lead to extremely hard negative examples.

\begin{table}[htb!]
  \centering
   \caption{The accuracy@k of the models on the considered data sets. The best results (bold) were achieved by Dummy and node2vec models.}
    \begin{tabular}{l|rrrr}
    acc@k & \multicolumn{1}{l}{Agda stdlib} & \multicolumn{1}{l}{Agda unimath} & \multicolumn{1}{l}{Agda TypeTopology} & \multicolumn{1}{l}{Lean Mathlib4} \\
    \midrule
    Dummy & 0.16  & 0.10  & 0.12  & \textbf{0.09 }\\
    BoW   & 0.00  & 0.02  & 0.00  & 0.00 \\
    TFIDF & 0.04  & 0.06  & 0.03  & 0.05 \\
    fastText & 0.00  & 0.05  & 0.02  & \multicolumn{1}{r}{NA} \\
    analogies & 0.04  & 0.03  & \multicolumn{1}{r}{NA} & \multicolumn{1}{r}{NA} \\
    node2vec & \textbf{0.29}  &\textbf{ 0.27}  & \textbf{0.15}  & 0.00 \\
    \end{tabular}%
  \label{tab:acc-k}%
\end{table}%

\begin{table}[htb!]
  \centering
  \caption{The precision (at $\vartheta = 0.5$) of the models on the considered data sets. The best results (bold) are obtained by various models, but only node2vec models have non-trivial recall as well.}
    \begin{tabular}{l|rrrr}
    precision & \multicolumn{1}{l}{Agda stdlib} & \multicolumn{1}{l}{Agda unimath} & \multicolumn{1}{l}{Agda TypeTopology} & \multicolumn{1}{l}{Lean Mathlib4} \\
    \midrule
    Dummy & 0.98  &\textbf{ 1.00}  &\textbf{ 1.00}  & \textbf{1.00} \\
    BoW   & 0.94  & 0.97  & 0.93  & 0.98 \\
    TFIDF & 0.98  & 0.99  & 0.99  & \textbf{1.00} \\
    fastText & 0.98  & 0.98  & 0.98  & NA \\
    analogies & \textbf{0.99}  & 0.99  & NA    & NA \\
    node2vec & 0.97  & 0.97  & 0.98  & 0.94 \\
    \end{tabular}%
  \label{tab:precision}%
\end{table}%

\begin{table}[htb!]
  \centering
  \caption{The recall (at $\vartheta = 0.5$) of the models on the considered data sets. The best results (bold) are obtained by node2vec models.}
    \begin{tabular}{l|rrrr}
    recall & \multicolumn{1}{l}{Agda stdlib} & \multicolumn{1}{l}{Agda unimath} & \multicolumn{1}{l}{Agda TypeTopology} & \multicolumn{1}{l}{Lean Mathlib4} \\
    \midrule
    Dummy & 0.01  & 0.07  & 0.06  & 0.02 \\
    BoW   & 0.00  & 0.01  & 0.00  & 0.00 \\
    TFIDF & 0.03  & 0.04  & 0.02  & 0.01 \\
    fastText & 0.02  & 0.03  & 0.01  & NA \\
    analogies & 0.03  & 0.02  & NA    & NA \\
    node2vec & \textbf{0.95}  &\textbf{ 0.94 } &\textbf{ 0.98}  & \textbf{0.97} \\
    \end{tabular}%
  \label{tab:recall}%
\end{table}%

\begin{table}[htb!]
  \centering
  \caption{The area under the ROC curve of the models on the considered data sets. The best results (bold) were achieved by node2vec models.}
    \begin{tabular}{l|rrrr}
    AU-ROC & \multicolumn{1}{l}{Agda stdlib} & \multicolumn{1}{l}{Agda unimath} & \multicolumn{1}{l}{Agda TypeTopology} & \multicolumn{1}{l}{Lean Mathlib4} \\
    \midrule
    Dummy & 0.51  & 0.54  & 0.53  & 0.51 \\
    BoW   & 0.50  & 0.51  & 0.501 & 0.50 \\
    TFIDF & 0.51  & 0.52  & 0.51  & 0.51 \\
    fastText & 0.51  & 0.52  & 0.50  & NA \\
    analogies & 0.51  & 0.51  & NA    & NA \\
    node2vec & \textbf{0.99 } &\textbf{ 0.98}  & \textbf{1.00 } &\textbf{ 0.99} \\
    \end{tabular}%
  \label{tab:aucroc}%
\end{table}%

\begin{figure}
    \centering
    \includegraphics[width=\textwidth]{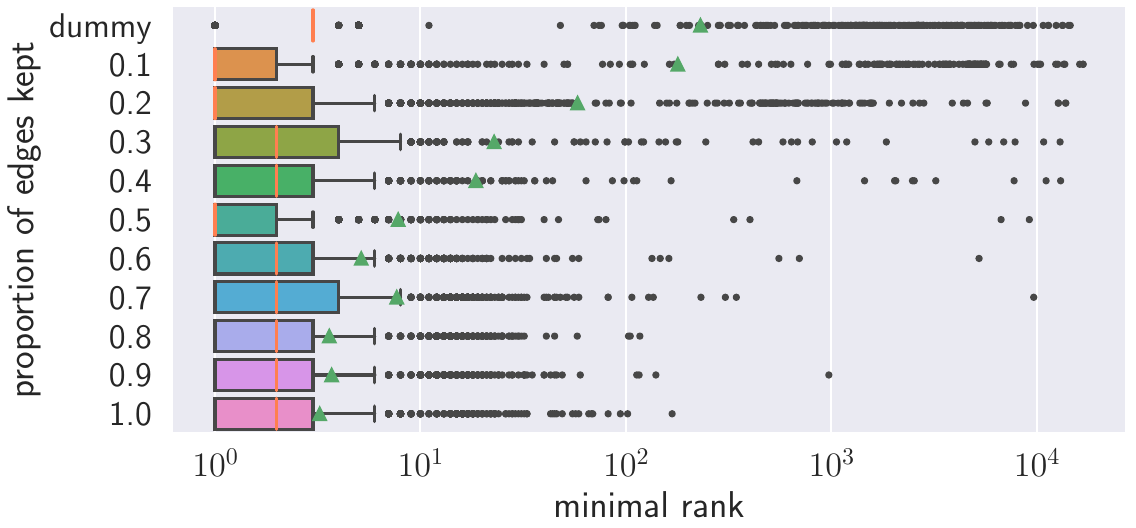}
    \caption{The distributions of the minimal ranks of the correct references for the test examples when varying the proportion of edges from the Agda stdlib library included in the training set from 0.1 and 1.0 (all edges). The box plot on the top corresponds to the Dummy recommendation system. The orange lines and the green triangles depict median and average minimal ranks, respectively.}
    \label{fig:boxplots}
\end{figure}

\paragraph{Ablation study.} 
We performed an ablation study of node2vec on the Agda standard library in the following way. We keep the test set intact to obtain comparable results and only manipulate the training set by keeping the proportion $p\in\{0.1, 0.2, \dots, 0.9, 1.0\}$ of the total weight of the edges in it. Since node2vec works in a transductive setting, all versions of the training set include all the nodes. To better understand the (aggregated) results, we show the distribution of minimal ranks for every training set as box plots in Figure~\ref{fig:boxplots}. In addition to the results of node2vec on manipulated training sets, the figure also contains the results of the Dummy model on the original training set.

We can observe a large variance in the minimal ranks produced by the Dummy model. This is explained by the fact that the Dummy model always assigns top ranks to the most referenced entries while the ranks of the other entries are randomly distributed. Therefore, the Dummy model performs very well on the test entries that reference the most referenced entries, while at the same time, it performs pretty badly on the other test entries.

Analyzing node2vec results, we can see that training the model on only $p = 10\%$ of the edges already leads to a better performance of node2vec as compared to the Dummy model (in terms of the average minRank). As the value of $p$ increases from $0.1$ to $1.0$, the median of the minRank does not change much, while the average minRank is getting lower since there are fewer test examples on which the model performs \emph{extremely} badly.
\color{black}

\subsection{Repository and GDrive folder structure}
The structure of the GitHub repository (preprocessing scripts and the code for learning) is described in its README file. In this document, we only describe the structure of the GDrive folder.

In the folder, every data set (library) resides in a separate directory with a name resembling the library name (e.g., \texttt{stdlib}). Each directory contains
\begin{itemize}
    \item the file \texttt{network.csv},
    \item a (zipped) directory \texttt{entries} with the \texttt{.dag} files.
\end{itemize}
The names of DAG files are two-part: the first part is the namespace to which a given entry belongs. This is the actual namespace of a Lean entry and the fully qualified module name of an Agda entry. The second part of the name is the entry number, which ranges from $0$ (included) to the number of entries in an s-expression (excluded). For example, one of the \texttt{stdlib} entries can be found in \texttt{stdlib/entries/Data.Fin.Properties\_0147.dag}.

All the data can be loaded by running \texttt{main.py} script. It needs two packages to run: \texttt{networkx} for storing the network and \texttt{tqdm} for showing the progress. They can be installed by issuing the command \texttt{pip install -r requirements.txt} (the requirements file is also present). The computational graphs are stored as members of tree-like class \texttt{Entry}, defined in \texttt{main.py}. The code was tested with Python 3.11. 

For convenience, we uploaded \texttt{entries.zip} to the temporary Google Drive location. When running \texttt{main.py} for the first time, the directory \texttt{entries} will be automatically created and populated with unzipping the \texttt{entries.zip} file.

\section{Intended uses}
\emph{MLFMF} data sets were created to support further improvement of the numerous machine learning approaches to formalized mathematics. Primarily, the data sets can be used to evaluate the efficiency of the recommendation systems used to support formalization of mathematics with proof assistants. These systems help humans identify which previous entries (theorems, constructions, datatypes, and postulates) are relevant in proving a new theorem or carrying out a new construction. Please refer to Sections~3.5 and Section~4 of the main article for further details.

However, the data set collection can also serve as an appropriate benchmark for machine learning from graphs. For example, the node types can be used as the target concept of node classification.

\section{Hosting, Maintenance and Access}
The data is currently available at the anonymous Google Drive folder \url{https://drive.google.com/drive/folders/1o6QP3Vo_9KR5gB8GPYKeSXDW8pPN8i2U?usp=sharing}. The link from the main text (\url{https://github.com/ul-fmf/mlfmf-data}) points to a repository that contains the rest of the code, as well as the README file.

After the reviewing process, the collection of the data sets will be published as is in the aforementioned (publicly available) GitHub repository under the Creative Commons Attribution 4.0 International\footnote{https://creativecommons.org/licenses/by/4.0/} (CC BY 4.0). 

Note that the data sets are based on the source code of Agda and Lean libraries that evolves through time (entries might get added, deleted, or modified), but we will obtain a persistent dereferenceable identifier for the current snapshot. Moreover, we plan to update each data set when the underlying library significantly changes. This is not a rare event, given that, e.g., the size of \texttt{unimath} library almost doubled in the last six months.





\section{Author Statement}
The authors bear all responsibility in case of violation of rights related to the source data, i.e., publicly and freely available libraries in Agda and Lean. The authors also bear all the responsibility associated with the eventual breach of the licenses of the data sources.